\documentclass[10pt,twocolumn,letterpaper]{article}

\usepackage{wacv}              % To produce the CAMERA-READY version
\usepackage{graphicx}
\usepackage{amsmath}
\usepackage{amssymb}
\usepackage{booktabs}
\usepackage{svg}
\usepackage{float}
\usepackage[accsupp]{axessibility}  % Improves PDF readability for those with disabilities.
\usepackage{algorithm}
\usepackage{algorithmic}
\usepackage[pagebackref,breaklinks,colorlinks]{hyperref}

\usepackage{epsfig}
\usepackage{tabularx}
\usepackage{multirow,bigdelim}
\usepackage{enumitem}
\usepackage{bm}

\usepackage[font=footnotesize,labelfont=bf]{caption}
\usepackage{pifont}% http://ctan.org/pkg/pifont
\newcommand{\MTTSortBody}{
\begin{figure*}[t]
  \centering
  \includegraphics[trim={0.01cm 0.01cm 0.01cm 0.01cm},clip,width=0.85\linewidth]{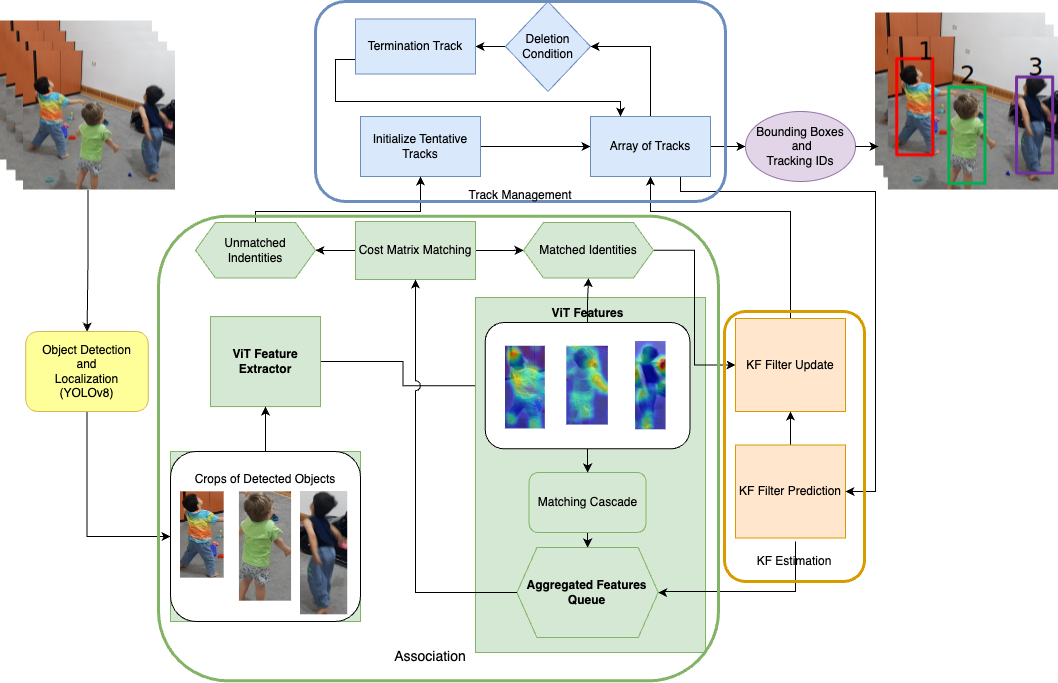}
  \caption{Our proposed method, MTTSort for multiple toddler tracking in indoor videos. This diagram illustrates two significant enhancements to the traditional DeepSort framework: (1) Pooled Aggregated Feature Association with a Custom Buffer, a mechanism that accumulates and consolidates features across consecutive frames in a user-defined buffer, and (2) Attention-Based Feature Extraction with ViT, which replaces conventional CNNs with the Vision Transformer for a more refined and attention-focused feature extraction process. Both modifications are designed to tackle the challenges posed by subjects like toddlers, characterized by their similar appearances and unpredictable movements.}
  \label{fig:mttsortbody}
\end{figure*}
}

\newcommand{\errosfig}{
\begin{figure}[ht!]
  \centering
  \includegraphics[width=0.70\columnwidth]{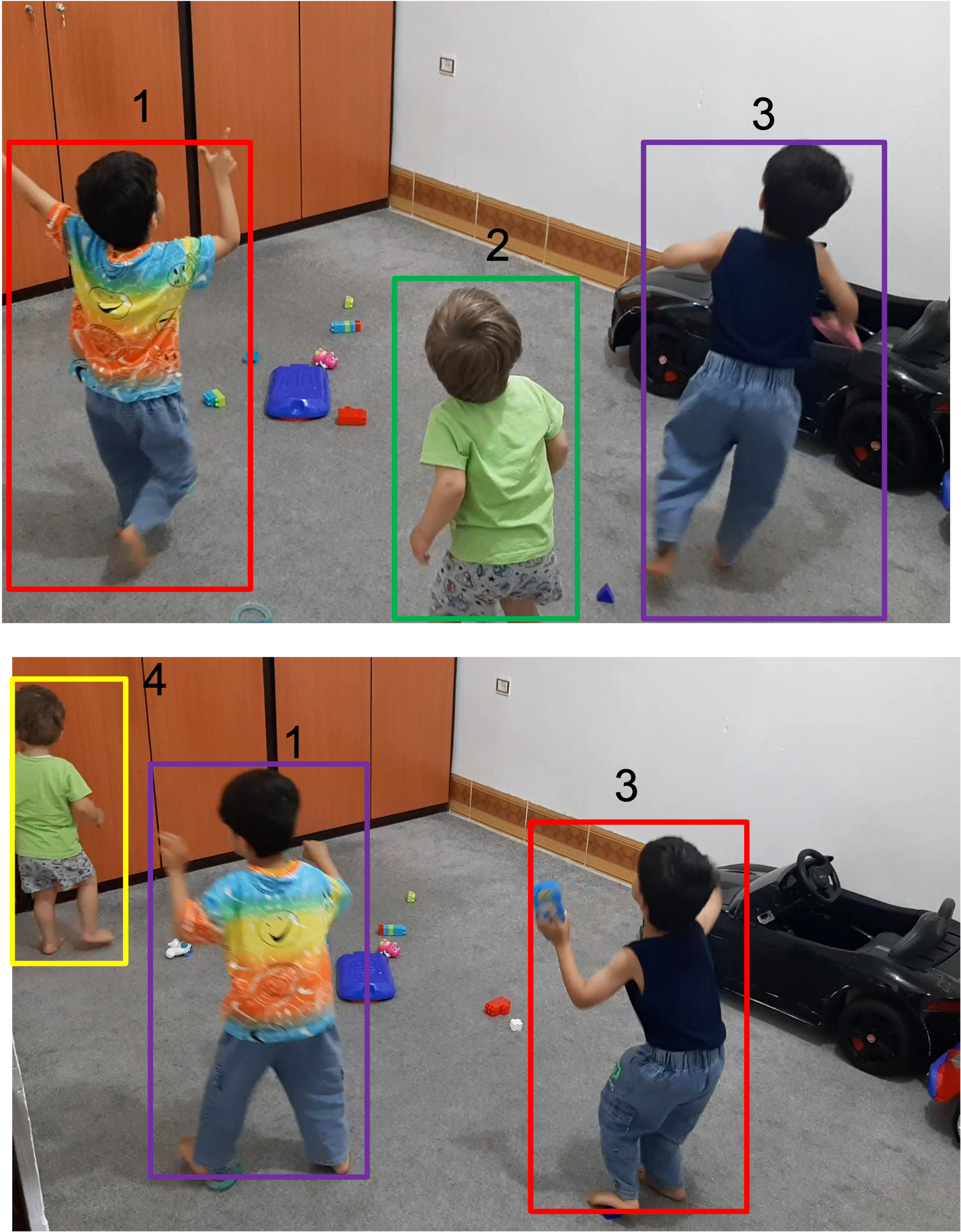}
  \caption{ID switch and fragmentation errors: toddler 1 and toddler 3 in the top image, have had their ID numbers swapped with each other in the bottom image, constituting an ID switch error. Toddler 2, present in the top image, is no longer tracked in the bottom image and is treated as a new toddler assigned the ID 4, indicative of a fragmentation error.}
  \label{fig:errofig}
\end{figure}
} 
\newcommand{\tabParametrs}{
\begin{table}[h]
    \centering
    \caption{MOT hyperparameters with descriptions.}

    \small % decrease font size
    \setlength{\tabcolsep}{2pt} % decrease column spacing
    \begin{tabular}{|c|c|p{3cm}|}
    \hline
    \textbf{Parameter} & \textbf{Range} & \textbf{Description} \\
    \hline
    \texttt{MAX\_DIST} & [0.1, 1.0] & Maximum distance for matching \\
    \hline
    \texttt{MIN\_CONFIDENCE} & [0.2, 0.7] & Minimum confidence for detection \\
    \hline
    \texttt{NMS\_MAX\_OVERLAP} & [0.2, 0.8] & Max overlap for non-max suppression \\
    \hline
    \texttt{MAX\_IOU\_DISTANCE} & [0.2, 0.9] & Max IoU distance for tracking \\
    \hline
    \texttt{MAX\_AGE} & [10, 200] & Max age of a track without detection \\
    \hline
    \texttt{N\_INIT} & [2, 15] & Number of initial detections \\
    \hline
    \texttt{NN\_BUDGET} & [20, 120] & Nearest neighbors' budget \\
    \hline
    \end{tabular}
    \label{tab:1_deepsort_parameters}
\end{table}
}
\newcommand{\tabConfigurations}{
\begin{table}[h!]
\centering
\caption{Accuracy parameters in different configurations. Each configuration is designed to address a specific challenge. Configuration 7 with parameters resulting from the proposed genetic algorithm has achieved the best accuracy.}
\label{table:2_conf}
\small
\setlength{\tabcolsep}{4pt}
\begin{tabular}{|c|c|c|c|c|c|}
\hline
\textbf{Config.} & \textbf{MOTA} & \textbf{DetRe} & \textbf{DetPr} & \textbf{DetA} & \textbf{HOTA} \\
\hline
1 & 0.94 & 0.96 & 0.92 & 0.89 & 0.56 \\
\hline
2 & 0.95 & \textbf{0.97} & 0.94 & 0.91 & 0.62 \\
\hline
3 & 0.94 & 0.94 & 0.94 & 0.89 & 0.58 \\
\hline
4 & 0.92 & 0.89 & 0.95 & 0.85 & 0.59 \\
\hline
5 & 0.92 & 0.89 & 0.95 & 0.85 & 0.66 \\
\hline
6 & 0.91 & 0.87 & 0.94 & 0.82 & 0.60 \\
\hline
\textbf{7} & \textbf{0.95} & 0.92 & \textbf{0.99} & \textbf{0.91} & \textbf{0.67} \\
\hline
\end{tabular}
\end{table}
}

\newcommand{\tabAlgorithmComparison}{
\begin{table*}[ht!]
\centering
\caption{A quantitative comparison between different algorithms on indoor and outdoor environments. The proposed method (MTTSort) has achieved the best results on the MTTrack dataset. DeepSort+GA shows a configuration of DeepSort resulting from the genetic algorithm. DeepSort+GA improves the HOTA and IDF1 on the MTTrack dataset significantly in comparison with the traditional DeepSort.}
\label{table:algorithm_comparison}
\footnotesize % Slightly larger than \small
\setlength{\tabcolsep}{3pt} % Slightly larger spacing between columns
\begin{tabular}{|l|ccc|ccc|ccc|}
\hline
\textbf{MOT Algorithms} & \multicolumn{3}{c|}{\textbf{Outdoor (MOT15)}} & \multicolumn{3}{c|}{\textbf{Indoor (DanceTracker)}} & \multicolumn{3}{c|}{\textbf{Indoor (MTTrack)}} \\
\cline{2-10}
& \textbf{MOTA} & \textbf{HOTA} & \textbf{IDF1} & \textbf{MOTA} & \textbf{HOTA} & \textbf{IDF1} & \textbf{MOTA} & \textbf{HOTA} & \textbf{IDF1} \\
\hline
DeepSort \cite{2DeepSort10011018} & \textbf{0.77} & 0.78 & 0.86 & 0.79 & 0.33 & 0.49 & 0.94 & 0.56 & 0.82 \\
\hline
StrongSort \cite{9_du2023strongsort} & 0.74 & 0.76 & 0.85 & 0.75 & 0.31 & 0.47 & 0.91 & 0.47 & 0.78 \\
\hline
HybridSort \cite{10_yang2023hybridsort}& \textbf{0.77} & \textbf{0.80} & \textbf{0.88} & \textbf{0.90} & \textbf{0.48} & \textbf{0.54} & 0.87 & 0.20 & 0.86 \\
\hline
Bytetrack \cite{11_BT10.1007/978-3-031-20047-2_1} & 0.76 & 0.70 & 0.80 & 0.80 & 0.47 & 0.52 & 0.96 & 0.52 & 0.96 \\
\hline
\textbf{DeepSort+ GA} & 0.69 & 0.57 & 0.62 & 0.65 & 0.39 & 0.22 & 0.95 & 0.67 & 0.97 \\
\hline
\textbf{MTTSort (Ours)} & 0.66 & 0.59 & 0.67 & 0.72 & 0.43 & 0.23 & \textbf{0.98} & \textbf{0.68} & \textbf{0.98} \\
\hline
\end{tabular}
    \label{tab:3_comp}

\end{table*}
}

\usepackage[capitalize]{cleveref}
\crefname{section}{Sec.}{Secs.}
\Crefname{section}{Section}{Sections}
\Crefname{table}{Table}{Tables}
\crefname{table}{Tab.}{Tabs.}

\def\wacvPaperID{3} % *** Enter the WACV Paper ID here
\def\confName{WACV}
\def\confYear{2024}

\begin{document}

\title{Multiple Toddler Tracking in Indoor Videos}
\author{Somaieh Amraee$^{1,2}$, Bishoy Galoaa$^1$, Matthew Goodwin$^3$, \\ Elaheh Hatamimajoumerd$^{1,2}$, Sarah Ostadabbas$^{1*}$ \\
$^1$Department of Electrical \& Computer Engineering, Northeastern University, MA, USA\\
$^2$ Roux Institute at Northeastern University, ME, USA\\
$^3$ Bouve College of Health Sciences, Northeastern University, MA, USA\\
{$^*$Corresponding author's email: \tt\small Ostadabbas@ece.neu.edu}
}
\maketitle

\begin{abstract}
Multiple toddler tracking (MTT) involves identifying and differentiating toddlers in video footage. While conventional multi-object tracking (MOT) algorithms are adept at tracking diverse objects, toddlers pose unique challenges due to their unpredictable movements, various poses, and similar appearance. Tracking toddlers in indoor environments introduces additional complexities such as occlusions and limited fields of view. In this paper, we address the challenges of MTT and propose MTTSort, a customized method built upon the DeepSort algorithm. MTTSort is designed to track multiple toddlers in indoor videos accurately. Our contributions include discussing the primary challenges in MTT, introducing a genetic algorithm to optimize hyperparameters, proposing an accurate tracking algorithm, and curating the MTTrack dataset using unbiased AI co-labeling techniques. We quantitatively compare MTTSort to state-of-the-art MOT methods on MTTrack, DanceTrack, and MOT15 datasets. In our evaluation, the proposed method outperformed other MOT methods, achieving 0.98, 0.68, and 0.98 in multiple object tracking accuracy (MOTA), higher order tracking accuracy (HOTA), and iterative and discriminative framework 1 (IDF1) metrics, respectively\footnote{The MTTSort code available at \href{https://github.com/ostadabbas/Multiple-Toddler-Tracking}{https://github.com/ostadabbas/Multiple-Toddler-Tracking}.}.

\end{abstract}

%%%%%%%%% BODY TEXT
\section{Introduction}
\label{sec:intro}
%\com{SO: I did some writing for the first three paragraphs but you need to add more citations.}

%\com{Pharag. 1 - SO: I did some writing but you need to add more citations.}
Multiple toddler tracking (MTT) involves the detection of toddlers in video footage and continuous tracking with a unique identification number. According to the American Academy of Pediatrics (AAP), toddlers are children aged 1-3 years characterized by their active engagement in activities such as climbing, running, and jumping \cite{1toddler}. Ensuring the safety of toddlers during their daily routines, both at home and in care facilities, is a paramount concern for parents and their assigned caregivers. The identification and monitoring of toddlers’ movements is of significant interest for researchers in child development \cite{27Posture_basedHuang_2023_CVPR, 36_Huang_2023_WACV}, nursing \cite{28nns10.1007/978-3-031-43895-0_55}, and various health-related fields \cite{25_baby10042719}, particularly early detection of motor-related abnormalities 
\cite{23_babymotions22030866, 24_motion_ASD10.3389/fpsyg.2023.1140731, 26_baby_motionGONG2022102308}. 

%\com{SA: Added new citations (applications)}

 %\com{Pharag. 2 About MOT. Start with: Multiple toddler tracking is a special category of multi-object tracking (MOT)}
Multiple toddler tracking falls under the specialized category of multiple object tracking (MOT) algorithms. MOT
is a crucial component of various scene-understanding tasks, including surveillance  \cite{30_sur10064240}, robotics \cite{31_rootic8692964,32_robotic9242337}, and autonomous  \cite{33_AVs23084024}. These algorithms track multiple moving objects while assigning a unique identifier to each \cite{2DeepSort10011018,3SROT7533003,4Meimetis2021RealtimeMO}. Typically, various objects are present in each frame and may belong to different classes. MOT algorithms are comprised of three key steps: detection (finding all objects in a frame), localization (determining detected object positions in a frame), and association (matching objects across frames to maintain consistent identifiers)  \cite{5app12031319}. Particularly in human tracking and monitoring systems, ensuring each person has unique and persistent identifier numbers throughout the video is a critical requirement.

%These systems are divided into two main categories: single-camera \cite{} and multi-camera \cite{} systems. In multi-camera systems, in addition to tracking the different human subjects, there is a need for different instants of each subject to match with others in different views. This means that in these systems, there are several independent tracks for each subject, which must have the same ID.  And it makes everything more difficult and requires more complex algorithms for tracking \cite{}.

%\com{Pharag. 3: The challenges for MTT}
While some toddler detection methods have been proposed to measure the distance of toddlers from dangerous objects and prevent potentially hazardous situations  \cite{1toddler,20_toddler_9194666}, it is evident that developing an algorithm capable of simultaneously tracking multiple children with unique identifiers is significantly more challenging. Multiple toddler tracking (MTT) presents three primary obstacles. First, many existing face and bounding box detection methods are trained primarily on adult samples, leading to numerous errors when applied to child detection (i.e., detection challenge) \cite{16wan2022infanface}. Second, young children exhibit unpredictable movement patterns involving rapidly changing directions and positions, such as walking, sitting, and crawling, making it challenging to establish an accurate tracking model (i.e., localization challenge) \cite{6_similarApp10030951}. Third, when multiple young children are present in a scene, distinguishing them can be problematic due to their similar appearances (i.e., association challenge) \cite{16wan2022infanface,6_similarApp10030951 }. Furthermore, to address the demand for intelligent systems for toddler tracking, they should be designed for indoor environments, such as homes and daily care centers. In these applications, the limitation of field of view, results in some challenges such as a high rate of occlusion \cite{8_indoor_9816269}. These challenges underscore the essential need for a customized and accurate method for tracking multiple toddlers in indoor videos.

%\com{Pharag. 4: Our Contributions}
This paper discusses the main challenges and potential solutions of MOT methods for multiple toddlers tracking in indoor videos. Then a customized and accurate method, MTTSort, is proposed to resolve these challenges and achieve high efficiency for toddler tracking. This method builds upon the DeepSort algorithm \cite{2DeepSort10011018} which is a state-of-the-art MOT approach that has demonstrated significant potential for customization in indoor applications in our experiments. In the first step of the proposed method, a genetic algorithm is proposed to optimize the hyperparameters. Then a new extension of the DeepSort algorithm is developed for multiple toddler tracking in indoor videos.  This paper also provides a quantitative comparison of state-of-the-art methods including DeepSort \cite{2DeepSort10011018}, StrongSort \cite{9_du2023strongsort}, HybridSort \cite{10_yang2023hybridsort}, and Bytetrack \cite{11_BT10.1007/978-3-031-20047-2_1}.  Since there is no publicly available dataset for multiple toddler tracking, we built a   video set, called the MTTrack dataset. Comprehensive evaluation and comparison have been conducted on MTTrack and two other public tracking datasets, DanceTrack \cite{7dance_9879192}, and MOT15 \cite{12_MOT15}. In summary, the paper introduces several significant contributions:
\begin{itemize}
\item Discussing the main challenges of MOT methods for multiple toddler tracking applications in indoor videos.

\item Providing a genetic algorithmic method to make sure that the optimum hyperparameters are used for tracking. 

\item Proposing an accurate tracking algorithm that is customized for multiple subject tracking in indoor videos.

\item Building and annotating the MTTrack dataset using the AI co-labeling techniques, ensuring no algorithmic biases.

\item  A quantitative comparison of state-of-the-art MOT methods on two public datasets as well as the MTTrack dataset.

\end{itemize}

Overall, these contributions enhance our understanding of multiple object tracking and provide valuable resources for further research in this domain. The rest of this paper is structured as follows: State-of-the-art MOT methods and their shortcomings for multiple toddler tracking in indoor videos are described in Section \ref{sec:related}. Then, in Section \ref{sec:methods}, the proposed customized method for multiple toddler tracking is described.  The experimental results are shown in Section \ref{sec:results}. Finally, Section \ref{sec:conclusion} is dedicated to conclusions and recommendations for future research.

\section{Related Works}
\label{sec:related}
In this section, the widely used existing MOT methods are briefly described, and then the main shortcomings of these methods for multiple toddler tracking especially in indoor videos are discussed.

%\com{talk about the SOTA of the MOT and why they are failing for MTT for each method. ALso, at the end, list the Publicly available datasets in this domain. }

\textbf{DeepSort} \cite{2DeepSort10011018} is designed to track multiple objects in real-time applications by combining object detection with deep neural networks, specifically using  YOLO \cite{21_yolov3} and  Deep Association Network (DAN) \cite{22Danwang2020dan}. DeepSort extends the original simple online and real-time tracking \cite{3SROT7533003} (SORT) algorithm by improving the re-identification of objects after an occlusion. In this method, the object detection module detects objects in each frame. YOLOv8 has been used as a deep object detection model in this paper. DeepSort uses the Kalman filter to estimate the state of each track (i.e., position, velocity, size, and age) and predict its future location. The Kalman filter is used in combination with the data association module to update the state of each track based on the detected objects in the current frame and predict the state of each track in the next frame. The track management module in DeepSort is responsible for updating the state of each track over time and removing tracks that are no longer valid or reliable. This module keeps track of each object's position, velocity, size, and age, and uses this information to predict the object's future location and update its state. In addition to updating the state of each track, the track management module also performs track maintenance tasks such as track initiation and track termination. While DeepSort exhibits the capability to track multiple moving objects effectively in straightforward scenarios where the objects are widely spaced and follow uncomplicated trajectories, it tends to produce numerous errors in more intricate situations, particularly in indoor video environments characterized by prolonged occlusions.

\textbf{StrongSort}\cite{9_du2023strongsort} revisits the classic DeepSort tracker with the appearance-free (AFLink) model and Gaussian-smoothed interpolation (GSI). StrongSort first uses an enhanced correlation coefficient maximization (ECC) to account for motion noise caused by movements. Then, a modified Kalman filter that emphasizes non-linear motions is used to calculate the weightings during each update across frames. Lastly, for object association, StrongSort directly includes the motion information in addition to appearance for more accurate tracking. While StrongSort generally outperforms DeepSort as a more accurate tracker when evaluated on publicly available outdoor datasets, our experimental findings indicated an unexpected outcome. Specifically, in our experiments, the performance of StrongSort was notably inferior to DeepSort, particularly in indoor video scenarios. Consequently, we introduced our tracking method, based on DeepSort, with the primary objective of achieving precise tracking of multiple toddlers within indoor videos.

\textbf{HybridSort}\cite{10_yang2023hybridsort} improves tracking when long-standing failure cases are caused by heavy occlusion and clustering. In this situation, strong cues such as spatial and appearance information become unreliable simultaneously. In this research, they demonstrated an important finding that previously overlooked weak cues, such as confidence state, height state, and velocity direction, can compensate for the limitations of strong cues. Then, they proposed HybridSort by introducing simple modeling for the newly incorporated weak cues and leveraging both strong and weak cues. The design effectively and efficiently resolves ambiguous matches generated by strong cues and significantly improves association performance. A critical limitation we observed in our experiments regarding HybridSort is its performance inconsistency. Specifically, when toddlers are in motion, this method yields highly satisfactory results. However, it encounters significant challenges when toddlers are either stationary or exhibiting minimal movement, leading to a notable increase in tracking errors.

\textbf{ByteTrack} \cite{11_BT10.1007/978-3-031-20047-2_1} has been proposed to fix missing predictions by using low-confidence candidates in association, achieving good performance by balancing the detection quality and tracking confidence. It focuses on associating almost every detection box, including low-score ones, to recover true objects and filter out background detections. ByteTrack associates every detection box and uses similarities between tracklets to reduce false positives and negatives for low-score detection bounding boxes. It proposes a second matching stage in which low-confident detections are associated with unmatched tracks from the first stage. The low-confident detections are not used to start new tracks, ensuring no ghost tracks from low-confident false positive detections are introduced. The authors of ByteTrack show that this two-stage matching (TSM) improves the tracking performance when integrated into various frameworks. In our initial experiments, ByteTrack showed promising results. However, it became clear that it had limited potential for customization and parameter adjustment, particularly when dealing with indoor videos, in contrast to the DeepSort algorithm. Consequently, we made the decision to enhance and tailor DeepSort with our modifications to create a highly accurate tracker for the purpose of tracking multiple toddlers in indoor video settings.

\textbf{Public Datasets} for multi-object tracking have been organized into two main resources, MOT \cite{12_MOT15}  and DanceTrack \cite{7dance_9879192}. The MOT  datasets are designed for the task of multiple object tracking. There are several variants of the dataset released each year, such as MOT15, MOT17, and MOT20.
DanceTrack is a large-scale multi-object tracking dataset for human tracking in occlusion, frequent crossover, uniform appearance, and diverse body gestures. It is proposed to emphasize the importance of motion analysis in multiple object tracking instead of mainly appearance-matching-based diagrams.

\section{Our MTTSort}
\label{sec:methods}
Here, we outline the primary difficulties faced by MOT techniques when tasked with tracking multiple toddlers in indoor video environments. To address these challenges, we introduce our novel approach. Our method (MTTSort), which builds upon the DeepSort algorithm, consists of two pivotal phases. In the initial phase, we utilize a genetic algorithm to determine the optimal parameters for DeepSort. This crucial step ensures the use of optimized parameters prior to any further adjustments. Following this, in the second phase, we implement specific modifications to DeepSort, thereby improving its precision in tracking multiple toddlers in indoor video scenarios.

\subsection{MTT Main Challenges}
Utilizing a conventional MOT method for tracking toddlers in indoor videos presents several noteworthy challenges, as outlined below:

\textbf{Adult-Centric Models:} Existing detection and tracking models have predominantly been trained on adult samples, resulting in a significant number of errors when detecting children.

\textbf{Unpredictable Movements:} Young children exhibit unpredictable and rapid changes in their movements and positions, making it challenging to develop an accurate tracking model for them.

\textbf{Activity Variability:} Toddlers engage in diverse activities, including walking, sitting, lying, and crawling, all within a single video sequence. This variability introduces higher error rates in both detection and tracking processes.

\textbf{Similar Appearances:} Distinguishing between toddlers can be problematic due to their similar physical appearances. This challenge becomes more pronounced, particularly when tracking twins, given their resemblance.

\textbf{Toy Confusion:} In scenarios where a child interacts with humanoid toys or action figures, the detection stage may mistakenly identify the toy as a real subject, leading to tracking errors.

\textbf{Limited Camera View:} Indoor video setups often employ stationary cameras with restricted fields of view, resulting in frequent occlusions, even with a small number of subjects being tracked.

\textbf{Extended Occlusion:} Unlike outdoor environments where occlusion might occur over a few frames, indoor scenarios involve prolonged occlusion extending over consecutive frames. This extended occlusion poses a significant challenge in maintaining individual subjects' identities.

\textbf{Data Challenges:} Developing a novel model for tracking multiple toddlers necessitates a substantial dataset for training. However, data collection and labeling for children's research are expensive, individualized, and subject to stringent privacy and classification laws. Capturing multiple video clips from various children can also be time-consuming due to their unpredictable movements.

In summary, the importance of ``SmallData'' is evident, emphasizing the challenge of obtaining sufficient labeled data for toddler tracking. Customizing existing methods is crucial to adapting to the unique demands and complexities associated with toddler tracking, enabling the development of more precise tracking solutions tailored to their distinctive characteristics and requirements.

% \PoseCTDArch
\subsection{Parameter Optimization Using Genetic Algorithm}
In the realm of optimization problems, the Genetic Algorithm (GA) stands out as a bio-inspired heuristic, rooted in the process of natural selection. GAs operate by simulating the process of evolution found in nature. Beginning with a population of potential solutions (analogous to individuals or organisms), the algorithm iteratively evaluates, selects, mates, and mutates these solutions. The key principle is that over successive generations, better and fitter solutions emerge, closely resembling the evolutionary concept of "survival of the fittest." The algorithm's capacity to explore a vast solution space by intelligently combining and modifying solutions makes it especially effective for problems where the optimal solution is elusive or computationally intensive to ascertain directly\cite{34_GA8862255}.

%\com{BG: Added an introduction for the GA as well as a citation}

In our pursuit of tackling the complexities of multi-toddler tracking, we embarked on a comprehensive exploration of various configurations encompassing diverse aspects of the detection and tracking challenges. Following a rigorous evaluation, we concluded that DeepSort emerged as the most suitable solution to meet our tracking needs. Nonetheless, optimizing performance necessitated a meticulous fine-tuning of DeepSort hyperparameters using a genetic algorithm \cite{17_mantau2022ga}. We pinpointed specific hyperparameters, each associated with a defined range derived from empirical insights and domain expertise. These critical hyperparameters are shown in \autoref{tab:1_deepsort_parameters}.

To discover the optimal values within these ranges, we adapted a mainstream genetic algorithm. The fitness function for our genetic algorithm was designed to maximize the aggregate score, \( Score \), defined as:

\begin{equation}
Score = HOTA + MOTA + IDF1,
\end{equation}
where, HOTA (higher order tracking accuracy) \cite{13_Hota}, MOTA (multiple object tracking accuracy) \cite{14_MOTA}, and IDF1 (iterative and discriminative framework 1) \cite{15_IDF1_10.1007/978-3-319-48881-3_2} are the three main accuracy criteria in MOT algorithms. Each metric was given an equal weight. This approach allows us to automatically and efficiently search the hyperparameter space, ultimately leading to enhanced tracking performance in our MTT system. Algorithm 1 shows the main body of the proposed genetic algorithm for optimizing the parameters.

\tabParametrs

%\com{EH: how about make this psudo code more specific, like in line 3 you can add your own termination criteria or name the parameters that you want to potimize}

%\com{BG: Added a more specific pseudo-code}

\begin{algorithm}
\caption{Genetic algorithm for hyperparameter optimization in MOT}
\begin{algorithmic}[1]
\STATE \textbf{Define} config\_template with hyperparameters of \autoref{tab:1_deepsort_parameters}.
\STATE \textbf{Initialize} a population of individuals using the config\_template.
\STATE \textbf{Evaluate} the fitness of each individual in the population using the fitness function:

$Score = HOTA + MOTA + IDF1$.
\WHILE{standard deviation of scores in the population is greater than tolerance OR maximum number of generations not reached}
    \STATE \textbf{Select} parents from the current population based on their scores.
    \STATE \textbf{Perform} crossover (recombination) on pairs of parents to produce offspring.
    \STATE \textbf{Mutate} offspring based on mutation rate.
    \STATE \textbf{Evaluate} the score of the offspring using the fitness function.
    \STATE \textbf{Replace} the current population with the offspring.
\ENDWHILE
\RETURN The solution (individual) with the best score from the population.
\end{algorithmic}
\label{alg:1_GA_MOT}
\end{algorithm}

\subsection{Indoor MTT}
\MTTSortBody

In the realm of tracking applications, particularly those involving subjects like toddlers with highly similar appearances and unpredictable movements, conventional tracking algorithms such as DeepSort often grapple with maintaining consistent identities. Recognizing these challenges, we have introduced significant modifications, integrating a state-of-the-art feature association mechanism into the DeepSort framework. \autoref{fig:mttsortbody} illustrates the essence of our proposed method. Our method enhances the DeepSort algorithm by adding two parts to it: (1) pooled aggregated feature association with custom buffer, and (2) attention-based feature extraction with the vision transformer (ViT).

{\textbf{Pooled aggregated feature association with custom buffer:}} While traditional tracking methods, including DeepSort, predominantly rely on the immediate features from the current frame, our approach takes a leap forward. We've introduced a custom-sized feature buffer that aggregates features over a series of frames. Specifically, in our experiment, the buffer size was fixed to store up to 5 features extracted per object. This choice of a buffer size of 5 was deliberate; we found that increasing the buffer size further led to an accumulation of more divergent features over time. As a result, the matching process could become less accurate, since features could deviate significantly from the object's most recent appearance. Hence, a size of 5 strikes a balance between retaining recent appearance information and ensuring effective feature matching.

The features buffer was designed as a queue, and the queuing mechanism operates under two distinct conditions: (1) When the Kalman filter terminates a track, leading to the removal of associated features, or (2) when the buffer reaches its full capacity, implying that 5 features have already been buffered. In such a scenario, the oldest feature is dequeued, ensuring that only the 5 "last seen" features are stored. Essentially, this buffer acts as a temporal sliding window for each object, capturing the most recent and relevant appearance data over time.

This buffer retains the "last seen" features for each subject across multiple frames, which, when subjected to an average pooling operation, produces aggregated features that capture the historical appearance nuances of each subject\cite{marin2023token}. This innovation not only ensures that the most recent and pertinent features are always in play but also amplifies the reliability of associations. By pooling these features, our algorithm achieves a holistic representation, adeptly handling transient appearance changes, momentary occlusions, or drastic appearance shifts — a marked enhancement over traditional methodologies.

%\com{BG: I added some details regarding the way the feature buffer queue}

\textbf{{Attention-based feature extraction with ViT:}} DeepSort, like many tracking algorithms, has conventionally leaned on CNNs for feature extraction. While CNNs have been instrumental in many computer vision tasks, in our experiments they occasionally missed the mark in scenarios demanding meticulous attention to minute details. Addressing this gap, we have integrated the Vision Transformer (ViT), an attention-centric model \cite{18_li2022attention}, supplanting the conventional CNN in DeepSort.
The ViT, renowned for its self-attention mechanisms, shines in pinpointing subtle differences by zeroing in on vital image regions. This capability is paramount for our toddler tracking application, ensuring that even the most nuanced appearance variations are meticulously captured, offering a richer and more detailed feature set for association.
%\textbf{{Integration with DeepSORT and Performance Metrics:}}

%\subsection{Hyper Parameter Tuning using GA}

%

\section{Experimentation Results}
\label{sec:results}
In the conducted experiments, multiple configuration setups were systematically assessed to understand the robustness and efficiency of the object tracking model. Each configuration was run across five different sub-scenes, generating individual results per sub-scene. To aggregate the results, the metrics from each configuration were averaged across all the sub-scenes, providing a comprehensive view of the model's performance under varying conditions. This approach ensures that the derived insights and the comparative analysis are based on consistent and averaged data points, mitigating the impact of outlier sub-scenes on the overall evaluation. 

The experimentation was structured around several focused scenarios, each emphasizing different aspects of the model's parameters. One configuration concentrated on detection confidence, another on distance measures, the third on overlapping and intersection over union (IoU), and the last on age and budget of the tracks. These configurations were crafted to observe the impact of selective variation of parameters on the model's outcomes and to deduce which parameters are crucial for optimizing performance. Interestingly, during the experimentation, it was observed that varying only the IoU or the confidence did not significantly alter the results, implying a degree of robustness in the model against these parameters. Most configurations yielded similar performance metrics, indicating that the model's effectiveness is less sensitive to alterations in IoU and confidence values. This insight is instrumental in understanding the inherent stability of the model and guides further refinement and tuning of the model parameters.

%In conclusion, the structured and meticulous approach to experimentation and the thoughtful aggregation of metrics have enabled a deeper understanding of the model's behavior and performance under different conditions and parameter settings. The insights gained from this process are pivotal for enhancing the model's accuracy and reliability in object tracking tasks.

\subsection{Evaluation Criteria}
Accurate evaluation of MOT algorithms has proven to be very difficult, because MOT is a complex task, requiring accurate detection, localization, and association over time.  Generally, there are five types of errors in an MOT method: 1- false negative or misses when ground truth exists but the prediction is missed, 2-false positive when tracker prediction exists for no ground truth tracker, 3- merge or ID switch when two or more object tracks are swapped, 4- deviation which measures the average distance between the predicted location of an object and its true location over time, and  5-  fragmentation which shows a track suddenly stops getting tracked but the ground truth track still exists. It causes a false increment of identifier numbers.  \autoref{fig:errofig} shows an example of ID switches and fragmentation, which are the most challenging errors, in MOT algorithms. In this figure, there is an ID switch between toddler 1 and toddler 3. Also, toddler 2 has gotten a new ID due to the fragmentation error.

Using the mentioned five types of errors, various evaluation metrics can be calculated.  In this paper, HOTA \cite{13_Hota} (higher order tracking accuracy) is considered as the primary metric. HOTA combines several sub-metrics that evaluate algorithms from different perspectives, providing a comprehensive assessment of algorithm performance. In addition to HOTA, we also include other well-established metrics, such as MOTA \cite{14_MOTA} (multiple object tracking accuracy) and IDF1 
 \cite{15_IDF1_10.1007/978-3-319-48881-3_2} (iterative and discriminative framework1). IDF1 reflects the association aspect of the tracker, while MOTA is primarily influenced by detection performance. However, HOTA  explicitly measures both types of metrics and combines detection and association in a balanced way. It can be used as a single unified metric for ranking trackers.   

\errosfig
	
\subsection{Building Our MTTrack Dataset}
The dataset building was a sophisticated and detailed endeavor, primarily concentrating on toddler videos, which necessitated the precise and accurate labeling of the selected frames. To facilitate this intricate procedure, an innovative labeling technique was utilized, allowing for the efficient auditing and refinement of labels generated by an established MOT algorithm. This negated the need to initiate labeling from scratch for each frame, thus optimizing the process. To assure the integrity of the dataset and try to cancel biases towards specific algorithms, two distinguished MOT algorithms, StrongSort and DeepSort, were incorporated. The calculated average between the bounding boxes generated by these algorithms yielded unbiased and consistent labels, adding to the robustness of the dataset.

The MTTrack Dataset consists of recorded videos capturing three toddlers engaged in various activities within a room. These toddlers, aged 2-4 years, can be observed performing actions such as jumping, walking, sitting, and playing with tablets and toys. We formatted the dataset frames into 10 subscenes, each comprising a maximum of 300 frames.  This methodological division was instrumental in eliminating sudden changes in scenery and mitigating extreme ID switches, thereby ensuring the consistency and reliability of the labels \cite{19_dou2023identity}. The auditing and validation of the labels were meticulously executed using the Labelme \cite{35_labelmerussell2008labelme}open-source tool, enabling the verification of each label's accuracy, relevance, and compliance with established standards. This comprehensive approach to labeling, while extensive, was imperative in establishing a reliable and credible foundation for subsequent research phases, yielding a dataset of unparalleled accuracy and reliability. 
 
 %The meticulous and detailed scrutiny ensured the attainment of valid, relevant, and superior quality labels, pivotal for the success and credibility of the research undertakings.

\subsection{Experimentation Configurations}
To ensure a consistent evaluation, 10 toddler video clips, each with 300 frames, were used in our experiments using the MTTrack dataset. Also, we applied different MOT methods on two public datasets: MO15, and DanceTrack. To have a comprehensive evaluation, various configurations were tested to examine object tracking algorithms. Each configuration is tailored to address specific challenges and requirements in object tracking, ranging from high precision and reliability to robustness against occlusions and appearance changes. A brief description of each configuration is listed below:

\paragraph{Configuration 1:}
This default configuration serves as a balanced setup suitable for general-purpose object-tracking scenarios.
\paragraph{Configuration 2:}
With a heightened \texttt{MIN\_CONFIDENCE} of 0.7, this configuration is optimized to minimize false positives by considering only high-confidence detections.
\paragraph{Configuration 3:}
This configuration, with an increased \texttt{MAX\_DIST} of 0.4 and \texttt{MAX\_AGE} of 80, is tailored for scenarios where objects may change appearance significantly or be temporarily occluded.
\paragraph{Configuration 4:}
The reduced \texttt{NMS\_MAX\_OVERLAP} and \texttt{MAX\_IOU\_DISTANCE} of 0.3 in this setup makes it suitable for tracking smaller or thin objects in densely populated scenes.
\paragraph{Configuration 5:}
By allowing more overlap between bounding boxes and between detections and trackers, this configuration is suitable for tracking larger or blob-like objects where overlap is expected.
\paragraph{Configuration 6:}
This configuration targets challenging conditions with substantial appearance changes or noise, allowing even low-confidence detections due to a \texttt{MIN\_CONFIDENCE} of 0.3 and a \texttt{MAX\_DIST} of 0.6.
 
\paragraph{Configuration 7:} This configuration uses optimized parameters resulting from the proposed genetic algorithm. The GA algorithm's configuration was number of generations is 50, the pop size is 10, the mutation rate is 0.1, and the crossover rate is 0.7.

\subsection{Quantitative Comparison }
To conduct a thorough assessment, we carried out a series of experiments from various perspectives. Firstly we examined different configurations of DeepSort and compared them with the configuration resulting from the proposed genetic algorithm. \autoref{table:2_conf} shows the results of this experiment in terms of HOTA (higher order tracking accuracy), DetRe (detection recall), DetPr (detection precision), DetA (detection accuracy), and MOTA (multiple object tracking ac-
curacy). As can be seen in this table, configuration 7 which is the result of parameter optimization of the genetic algorithm achieved the best results.

\tabConfigurations

\tabAlgorithmComparison

\autoref{tab:3_comp} displays the results of benchmarked techniques when compared to our proposed method. Within this table, "DeepSort+GA" denotes an enhanced DeepSort configuration incorporating the proposed genetic algorithm. As depicted in the table, our method has delivered superior performance, particularly in the context of indoor videos, specifically excelling in the tracking of multiple toddlers, as evident in the MTTrack dataset. We also found out that the best performing parameters without the aggregated features were different from the ones we got after adding the ViT attention model and the aggregated features. Especially, the \texttt{N\_INIT} which decreased, suggesting that tracks are now initialized with fewer frames. This could be because the new features provide more distinct and reliable information early on.

In our experimental observations of HybridSort, a noteworthy drawback we encountered is its inconsistency in performance. As it can be seen in \autoref{tab:3_comp}, this approach delivers exceptionally good results when toddlers are active and moving. However, it faces considerable difficulties when toddlers remain still or display minimal movement, resulting in tracking errors. Based on the experimental results and the flexibility for customization and parameter adjustment offered by various MOT methods, we have determined to adopt the DeepSort algorithm as the baseline framework for our proposed method designed for multiple toddler tracking in indoor video footage.

%\deepsortplot

% \com { SA: @Bishoy please complete the last row of table2 }

\section{Conclusion}
\label{sec:conclusion}
This paper discussed the primary challenges of multiple object tracking methods for tracking toddlers in indoor videos. We then introduced a new tracking method named "MTTSort," which is designed for multiple toddler tracking. In the initial phase of MTTSort, we employed a genetic algorithm to estimate optimized tracking parameters. By melding our custom feature buffer and the ViT-based feature extraction, we've re-engineered the foundational components of the DeepSort algorithm in order to capture the temporal features of each subject. This rejuvenated algorithm underwent rigorous evaluation and benchmarked against performance metrics like MOTA, HOTA, and IDF1 on the collected MTTrack dataset and two public tracking datasets, MOT15, and DanceTrack. 

Looking ahead, our research will focus on addressing additional challenges in multiple toddler tracking, including scenarios involving action figures, crawling subjects, and twins detection and tracking. Furthermore, we plan to expand our research to encompass multi-view videos. This expansion will include work on multi-camera tracking and re-identification methods. Our ultimate goal is to implement our method in real-world multi-camera systems for tasks such as detection, tracking, action recognition of toddlers, and the prediction of potentially hazardous events in indoor videos.
{\small
\bibliographystyle{ieee_fullname.bst}
\bibliography{refs}
}

\end{document}